\documentclass[a4paper,UKenglish,cleveref, autoref, thm-restate]{lipics-v2021}

\usepackage{booktabs}

\usepackage{ulem}

\usepackage{color, colortbl}
\definecolor{Gray}{gray}{0.9}

\title{From meaning to perception - exploring the space between word and odor perception embeddings} 

\titlerunning{From meaning to perception} 

\author{Janek Amann\footnote{Corresponding author}}{University of Copenhagen, [Emil Holms Kanal, 2], Copenhagen, Denmark \and  }{ja@developdiverse.com}{}{}

\author{Manex Agirrezabal}{University of Copenhagen, [Emil Holms Kanal, 2], Copenhagen, Denmark \and \url{manexagirrezabal.github.io}}{manex.aguirrezabal@hum.ku.dk}{[0000-0001-5909-2745]}{}

\authorrunning{J.\, Amann and M.\, Agirrezabal} 

\Copyright{Janek Amann and Manex Agirrezabal} 



\category{} 

\relatedversion{} 

\nolinenumbers 

\hideLIPIcs  

\EventEditors{John Q. Open and Joan R. Access}
\EventNoEds{2}
\EventLongTitle{42nd Conference on Very Important Topics (CVIT 2016)}
\EventShortTitle{CVIT 2016}
\EventAcronym{CVIT}
\EventYear{2016}
\EventDate{December 24--27, 2016}
\EventLocation{Little Whinging, United Kingdom}
\EventLogo{}
\SeriesVolume{42}
\ArticleNo{23}

\begin{document}

\maketitle

\begin{abstract}
In this paper we propose the use of the \textit{Word2vec} algorithm in order to obtain odor perception embeddings (or smell embeddings), only using publicly available perfume descriptions. Besides showing meaningful similarity relationships among each other, these embeddings also demonstrate to possess some shared information with their respective word embeddings. The meaningfulness of these embeddings suggests that aesthetics might provide enough constraints for using algorithms motivated by distributional semantics on non-randomly combined data. Furthermore, they provide possibilities for new ways of classifying odors and analyzing perfumes.
We have also employed the embeddings in an attempt to understand the aesthetic nature of perfumes, based on the difference between real and randomly generated perfumes.
In an additional tentative experiment we explore the possibility of a mapping between the word embedding space and the odor perception embedding space by fitting a regressor on the shared vocabulary and then predict the odor perception embeddings of words without an a priori associated smell, such as \textit{night} or \textit{sky}.
\end{abstract}

\section{Introduction}

Imagine you're smelling a new perfume and want to describe its scent. Most likely you will refer to notes such as \textit{lemon}, \textit{mandarine} or \textit{musk}, that make up the complex scent of a perfume. Our goal in this work is to get vectorial representations (or embeddings\footnote{In this paper these will be referred to as \textit{smell embeddings}, \textit{note embeddings} or \textit{odor perception embeddings}}) for those notes that would capture some relevant properties. We would like these representations to capture that \textit{lemon} and \textit{mandarine} are more similar than lemon and musk. In this work we have created such representations using perfume descriptions\textemdash{}each perfume being represented by a list of notes.

But how do we know that the embeddings’ quality is good? The embeddings seem to capture some relevant relationships, as the model finds that e.g. citric components, such as, lemon, mandarine and so on, are relatively similar. Additionally, in order to have a systematic evaluation, we show that the information in our embeddings can be, to some extent, grounded in language, by comparing the relative distances among notes found in our embeddings and in pretrained word embeddings. Furthermore, we expand on this by demonstrating that the measured amount of shared information correlates with how much a word is associated with smell\footnote{In this paper the terms \textit{odor}, \textit{smell} and \textit{scent} will be used interchangeably. }. 

In addition, we tried to find the relation between the semantic information in word embeddings and the perceptual information in odor embeddings. We do that by fitting a regressor that maps one embedding space onto another embedding space. Using this regressor, the model is able to suggest a odor perception embedding for any word in the given language. 

These notes or perfume components can easily be associated with specific smells. Because of that, we tentatively believe that our model can  capture information in the intersection of smell, a sensory modality, and language. This might open up new ways of investigating how our brains form associations between semantic information and sensory experience.


The goal of this work is to create meaningful smell embeddings and to systematically validate them by using a ranking-based approach. Furthermore, we have employed the embeddings in an attempt to understand the aesthetic nature of perfumes, by analyzing the difference between real perfumes and randomly generated ones.
We contribute to the scientific community by providing a collection of odor perception embeddings, which are ready to be used by anyone\footnote{All our code and embeddings will be made available as a GitHub repository.}. We further believe that our findings regarding the composition of perfumes can be extrapolated to other areas where elements are combined in ways that are supposed to provide aesthetic value, such as harmony in music or flavor profiles in cooking.


The article is structured as follows. We start by providing the reader with some background information about perfumes. After that, we discuss the state of the art, in which we introduce works that have tackled the problem of representing perfumes, notes or molecules using different approaches. Then, we present the data and methods that we have employed in our experiments, which are introduced in the next section. We present the results for each of the experiments and discuss those outcomes in the discussion section. As an additional experiment, we present our explorations on mappings between vector spaces within the discussion section. Finally, we close with a conclusion and outline possible future directions.

\section{Background}

The scent of a perfume is usually classified in scent family and scent family sub types. Floral, Chypre, Fougère, Marine/Ozonic, Oriental, Citrus, Green, and Gourmand are considered to be the main family types. Examples for family sub types are  fruity, spicy, woody, or animalic  \cite{herz2011a}. In practice, however, there is some variation in what is considered a scent family. 
\cite{wise2000a} point out that grouping odors based on their qualitative similarity bears a huge potential for subjectivity, as basic sensory similarities and categories can be immensely influenced by a person's background and conception of reality.
Furthermore, based on \cite{chastrette1988a}, \cite{wise2000a} elaborate that, for instance, floral notes might be grouped together not because of a similar smell but rather because of the fact that they all come from flowers.
Based on this, such classifications cannot be viewed as a "rendition of truth, but of someone’s rendition of truth" \cite{wise2000a}. 
Aside from its family, a scent can further be described in terms of its \textit{notes}. A given combination of ingredients is referred to as a \textit{composition}.

A perfume composition consists of three types of notes: the \textit{top notes}, the \textit{heart notes}, the \textit{base notes}. All of these unfold in different ways over time. 
The top notes comprise of light, small and highly volatile molecules that evaporate rapidly. The heart notes come to light just before the top notes disappear, which can be anywhere between two minutes up to one hour after a perfume is applied. 
Base notes emerge when the heart notes disappear. They consist of heavy, large and relatively low-volatile molecules and are often fixatives that support the top and heart notes. They appear only after 30 minutes and can stay up to 24 hours. Given the dynamic nature of a perfume, its perceived smell changes over time \cite{herz2011a}\footnote{In order to leverage a shared vocabulary with pretrained word embeddings, we made no distinction between e.g. \textit{lemon} as a top note and \textit{lemon} as a heart note}. The perceived notes of a perfume, however, do not necessarily go back to an ingredient of the same name, e.g. "concrete". Those notes are commonly referred to as \textit{fantasy notes}. Please find below an example which includes the descriptors regarding the perfume \textit{Chance Eau Tendre} from Chanel.\footnote{\url{https://en.wikipedia.org/wiki/Chanel_Chance}}

\begin{itemize}
\item Top notes: Grapefruit and quince
\item Heart notes: Iris absolute and hyacinth
\item Base notes: Amber, white musk and cedar wood
\end{itemize}

The notes assigned to a perfume are semantic descriptors. Although widely applied, there are several questions surrounding these semantic descriptors, e.g. how many descriptors or references are needed in order to appropriately describe a perfume, or what is the right balance of scent families among the descriptors. \cite{wise2000a} also recognize the subjectivity involved in this method, since individuals have different olfactory experiences and might use different numbers of descriptors in order to describe an odor. However, they also add that averaging over several subjects provides some reliability.
The weak link between smell and language is also found in works \cite{majid2018differential,olofsson2015muted,yeshurun2010odor}. For example, in \cite{majid2018differential}, the authors found that when describing perceptual stimuli across languages, people tend to rely heavily on source-based descriptors.
Contrary to the widespread assumption that languages are universally insufficient at describing smell (e.g. \cite{herz2011a}), \cite{majid2014odors} show in a cross-lingual experiment that this might be true for English, but not necessarily for all languages. With regards to English, \cite{majid2020human} point out that smell-related language is quite infrequent compared to other sensory modalities such as vision.

Despite this apparent gap, \cite{IATROPOULOS201837} show how the distributional hypothesis and large text corpora can be leveraged to quantify the association between words and odors. 
Based on similar theoretical assumptions, \cite{gutierrez2018predicting} show that word embeddings trained on general text can be used as features for accurately predicting perceptual ratings of odors.

\section{State of the art}
There have been several attempts to analyze the representation of smell in embedded vectorial spaces. Some of them make use of Natural Language Processing (NLP) resources, others make use of odor characters as perceived by people, and some others combine them all.

In previous work analyzing the odor perception space and the use of language \cite{IATROPOULOS201837}, the authors introduce two metrics to describe how strongly a word is associated with olfaction and how specific a word is in its description of odors. Furthermore, they introduce a two-dimensional space that characterizes perceptual olfactory connotations of English-language odor descriptors.
We opted for a similar approach but decided to use a different evaluation procedure, which we describe in section 5.
The work by Hörberg et al. \cite{horberg2020mapping} is also rather related to ours. They propose a data-driven approach that automatically identifies odor descriptors in English, and then derive their semantic organization on the basis of their distributions in natural texts.
Other authors \cite{zarzo2020multivariate} analyze perfume notes (making a distinction between top, heart and base notes) and extract the Principal Components. They propose a visualization which is a modified version of the Hexagon of Fragrance Families, a sensory map that visualizes perfumes in a polygon. Each side of the polygon represents odor characteristics.

There are a number of works that make an attempt to predict odor descriptors based on molecules. A number of these works were presented as part of the DREAM olfaction prediction challenge \cite{Keller820} or make use of the publicly available data set \cite{keller2016olfactory}.
The authors in \cite{sanchez2019machine,sanchezchemistry} trained models to predict the odor of molecules. The problem, called quantitative structure-odor relationship (QSOR) modeling, is tackled using graph neural networks.
Khan et al. \cite{khan2007predicting} built an olfactory perceptual space and a molecular physicochemical space. They predict pleasantness of different molecules based on a Principal Component Analysis (PCA) reduction of the character space defined by Dravnieks \cite{dravnieks1985a}.
Other researchers \cite{nozaki2018predictive,gutierrez2018predicting} made use of vectorial representations of words for molecule-based odor prediction. 
While \cite{nozaki2018predictive} used Word2Vec embeddings \cite{mikolov2013b}, \cite{gutierrez2018predicting} made use of FastText word embeddings \cite{bojanowski2017enriching}.

\section{Data}

As suggested by researchers in \cite{goodwin2017a}, the website \textit{Basenotes}\footnote{\url{www.basenotes.net}} provides data for thousands of commercially available perfumes.
For this study, 42773 perfumes were obtained from the website. The top notes, heart notes and base notes of each perfume were collected\footnote{We do not have any explicit information about the number of authors involved. However, we have no reason to assume it was only one.}. We excluded perfumes with less than 3 listed notes or missing note category labels, which reduced the number of perfumes to 26253.
After extraction the lists of notes were lower-cased and all punctuation was removed (e.g. in order to match \textit{Ylang-ylang} and \textit{ylang-ylang}). We applied no further normalization processes. Consequently, notes such as \textit{sicilian lemon} and \textit{lemon} were treated as distinct notes. 
The final dataset consisted of 12550 notes distributed over 26253 perfumes with average number of notes $\mu = 8.6$ ($\sigma = 4.1$). Table 1 shows the 10 most frequent notes.

\begin{table}[]
    \centering
    \begin{tabular}{c|c}
    note   &  frequency \\
    \hline
    musk & 8601 \\
     bergamot & 7617 \\
     sandalwood & 7298 \\
     jasmine &  6426 \\
     amber & 6331 \\
     vanilla & 6322 \\
     patchouli & 6243 \\
     rose & 5166 \\
     cedarwood & 4074 \\
     vetiver & 3931 \\

    \end{tabular}
    \caption{The 10 most frequent notes and their raw counts}
    \label{tab:notes_rbos}
\end{table}

\section{Methods}

\subsection*{Word2vec}

For this study we decided to represent the notes of the perfumes as dense vectors (embeddings).
According to (\cite{jurafsky2019a}) dense embeddings outperform sparse representations in every NLP task. The assumed reasons for this are: 1) embeddings usually are of lower dimensionality and therefore a model has to learn fewer parameters, 2) the model might generalize better due to fewer parameters and 3) dense embeddings might be able to capture synonymity to a higher degree.

While there are several methods to obtain word embeddings, the method used in this study is called \textit{Word2vec} (\cite{mikolov2013a}; \cite{mikolov2013b}).
The \textit{Word2vec} method comprises two complementary approaches: the continuous bag-of-words model (CBOW) and the skip-gram model. The difference lies in the task that is used to obtain the representations. In the CBOW model, a word is predicted based on its context, whereas in the skip-gram approach it works the other way around.
The intuition behind both approaches is that the learned weights of the model used for prediction can be used as an embedding of the word. One of the great advantages of this is that it uses running text as implicitly supervised training data and therefore does not need manual labeling (\cite{jurafsky2019a}). This study uses the CBOW model, as it is the default architecture (\cite{rehurek_lrec}). Also, other hyperparameters such as the context window size were left at default, since assumptions about these hyperparameters are outside the scope of this study. They will, however, be subject of future works.

\subsection*{Rank Biased Overlap}
As we mentioned earlier, thanks to our perceptional embeddings, we were able to create rankings of the most similar notes given one specific note (or position in the embedded space). For example, by looking for notes similar to \textit{lemon}, we were able to find notes like \textit{bergamot}, \textit{orange} and so on. The notes’ ranking seemed to make sense based on our
linguistic/semantic intuition. In order to measure this intuition in a metric, we decided to use a metric that compares two similarity rankings (smell embedding space/word embedding space) and returns a measurement of how overlapping the two rankings are.

We employed Rank Biased Overlap (RBO) \cite{webber2010similarity}. RBO is a similarity metric for ranked lists and it can estimate a similarity measure between two rankings, even when not all elements in the rankings are exactly the same. Since RBO is a ranked metric, higher ranked overlaps can be set to have a higher weight, than lower ranks. The output of this metric is a value between $0$ and $1$. As a result, dissimilar rankings would return a value close to zero, while similar rankings would return a value close to one.


\section{Experiments}

This section covers the two main experiments of our study.

\subsection{Experiment 1a}

The goal of this experiment was to investigate whether the \textit{Word2vec} algorithm can be applied to perfumes similarly to language and create meaningful embeddings of the perfume notes. Although the perfume notes do not follow a strict sequential order the way language does, the word2vec algorithm can still be applied, since it processes the elements in the context window as bag of words. In order to simplify the process and make sure the choice of the context size would not impact the outcome too much, we represented each perfume as a sequence of 100 elements randomly sampled from its set of notes. We then trained the embeddings with 10, 20, 50 and 100 dimensions based on these sequences using the gensim implementation of the \textit{Word2vec} model \cite{rehurek_lrec}, leaving all parameters at default. 
In order to validate these smell embeddings we extracted the shared vocabulary between the smell embeddings and word embeddings (660 items). Our assumption was that although the two types of embeddings captured different information, there had to be some olfactory information in the word embeddings. This overlap could be quantified by comparing relative similarities, meaning if "lemon" and "orange" are close in word embeddings space they should also be relatively close in the smell embedding space\footnote{Another way of conceptualizing this is to think about the pretrained word embeddings capturing general language as opposed to the smell embeddings capturing exclusively olfactory language. With olfactory content also being present in general language (even if limited), there should be an intersection of the two.}. In order to test this assumption for every item in the shared vocabulary we ranked the remaining items by their cosine similarity both in the word embedding and the smell embedding space. We then compared the similarity rankings for every item by calculating the RBO. By this we received a value indicating agreement between the relative similarities for every item in the shared vocabulary. To see these values in perspective we repeated the procedure and randomly shuffled the items in the smell embedding space. Finally, in order to ensure that the measured agreement was not due to randomness, we performed a Mann-Whitney-U test on the two distributions of RBO-values. We repeated the same procedure with word embeddings trained on three different corpora\footnote{We used the unlemmatized continous skip-gram embeddings trained on English Wikipedia, Gigaword and Wikipedia+gigaword publicly available on \url{http://vectors.nlpl.eu/repository/}.} for all four odor perception embeddings in a grid search.
Due to the fact that olfaction is not too frequently expressed in English and that the word embeddings do not account for polysemy, we expected a significant but not necessarily striking overlap of the information expressed in the respective embeddings.

\subsubsection*{Results}

\begin{table}[]
    \centering
    \begin{tabular}{l|c|c|c|c}
    smell embedding size   &  word embedding & mean RBO & mean RBO random & p-value\\
    \hline
    10 & giga & .046 & .013 & <.001*** \\
    10 & wiki & .042 & .018 & <.001*** \\
    10 & wiki+giga & .047 & .016 & <.001*** \\
    \hline
    20 & giga & .049 & .013 & <.001*** \\
    20 & wiki & .049 & .018 & <.001*** \\
\rowcolor{Gray}
    20 & wiki+giga & \textbf{.05} & .016 & <.001*** \\
    \hline
    50 & giga & .039 & .013 & <.001*** \\
    50 & wiki & .038 & .018 & <.001*** \\
    50 & wiki+giga & .039 & .016 & <.001*** \\
    \hline
    100 & giga & .03 & .013 & <.001*** \\
    100 & wiki & .031 & .018 & <.001*** \\
    100 & wiki+giga & .031 & .016 & <.001*** \\
    \end{tabular}
    \caption{Results of the grid-search in Experiment 1a. The best result is marked in bold font but also with a light gray background.}
    \label{tab:notes_rbos}
\end{table}

The results of the grid-search are shown in Table \ref{tab:notes_rbos}. Although the mean RBOs are quite low, the RBOs are in all configurations significantly different from the randomized ones. The highest mean RBO was obtained by the smell embeddings of size 20 and the word embeddings from the wiki+giga corpus.

\begin{table}[]
    \centering
    \begin{tabular}{l|c}
    note  &  5 most similar notes \\
    \hline
    musk & amber, jasmine, sandalwood, bergamot, mandarin \\
    bergamot & lemon, mandarin, jasmine, petitgrain, rosemary \\
    sandalwood & jasmine, amber, musk, patchouli, freesia \\
    \hline
    coffee & roasted cocoa bean, tiramisone, dry ambered cherries, cacao \\
    lavender & lavandin, geranium, basil, menthol, rosemary \\
    vanilla & heliotrope, praline, almond blossom, tonka bean, sandalwood \\
    rosemary & lavender, bergamot, lemon, oregano, lavander \\
    strawberry & balsamic vanilla, cosmopolitan cocktail accord, creamy caramel, raspberry, peach \\
    incense & myrrh, papyrus, noble laurel, black tolu, opponax \\

    \end{tabular}
    \caption{The upper row shows the 5 most similar notes to the 3 most frequent one. The lower row shows the 5 most smilar notes to several selected ones.}
    \label{tab:most_five_similar}
\end{table}

Leveraging one of the known properties of \textit{Word2vec}, we took a closer look at some of the similarities between the smell embeddings. Table 3 shows some notes with their respective 5 most similar ones. The upper part of the table shows the three most common notes. There are two things that can be observed here: Firstly, among the three most common and their respective most similar notes, there seems to be a notable overlap. However, frequency does not seem to be the only factor involved, since the rankings seem to reflect semantic properties. \textit{Musk} and \textit{amber} are both considered animalic smells\footnote{https://www.fragrantica.com/notes/}. Interestingly, \textit{jasmine}, commonly considered a floral scent, is also said to have musky qualities\footnote{https://perfumesociety.org/ingredients-post/jasmine-2/}. This effect shows even more in the case of \textit{bergamot}. Here, the three out of 5 most similar notes(\textit{lemon}, \textit{mandarin} and \textit{petitgrain}) are citric as well. There might also exist subtle perceptual similarities with the notes \textit{jasmine} or \textit{rosemary}. But again, a frequency effect seems rather obvious. 
Looking at the lower part of the table, one can find several examples that make intuitive sense, \textit{lavender} or \textit{rosemary }being quite impressive examples, the latter even capturing the different spelling of \textit{lavander}. Let us look at the examples of \textit{coffee} and \textit{incense}: In the case of \textit{coffee}, the similarities to \textit{roasted cocoa bean} and \textit{cacao} are quite straight forward. Less straightforward but nonetheless fitting is the similarity to the synthetic note \textit{tiramisone}, which has a chocolate-like character \footnote{https://www.premiumbeautynews.com/en/symrise-is-enhancing-its-fragrance,13817}. For \textit{incense} we can see quite intuitive similarities to \textit{myrrh} and \textit{opponax}\footnote{Google reveals that the actual spelling should most likely be "opoponax"}, also known as \textit{sweet myrrh}, as well as to other spicy or woody notes like \textit{papyrus}, \textit{noble laurel} and \textit{black tolu}.

\subsection{Experiment 1b}

In order to add a second step to our systematic evaluation of the smell embeddings, we tested our assumption that the amount of agreement between the relative similarities should be correlated with how strongly a word is associated with olfaction. In a conceptually similar approach to \cite{IATROPOULOS201837}, we quantified the olfactory association of a word by computing its cosine similarity with the average vector of several olfaction-related words, such as "smell" or "odor"\footnote{The full list will be made available on GitHub}. We then computed Spearman's rank correlation coefficient for the RBO-values and our values for olfactory association. 

\subsubsection*{Results}

Even though the Spearman's correlation coefficient is rather small,
it does not seem to be result of a random process, based on the p-value obtained ($r = 0.16$ and $p<.001$). These results show that there is indeed a positive correlation between the RBOs and the olfactory association.

\subsection{Experiment 2}

The goal of this experiment was to assess if the smell embeddings can be used in order to investigate the formal aesthetics of perfumes. Inspired by \cite{birkhoff2013aesthetic} we made the following assumption: The notes of pleasant perfumes spread across the embedding spaces in specific ways, following some kind of order. While there exists a multitude of possible geometrical properties this assumption could manifest itself in, for the sake of keeping it simple, we hypothesized that in a pleasant perfume the individual notes are spread rather evenly around their centroid.

\begin{figure}
    \centering
    \subfloat[Histogram]{{\includegraphics[width=0.4\textwidth]{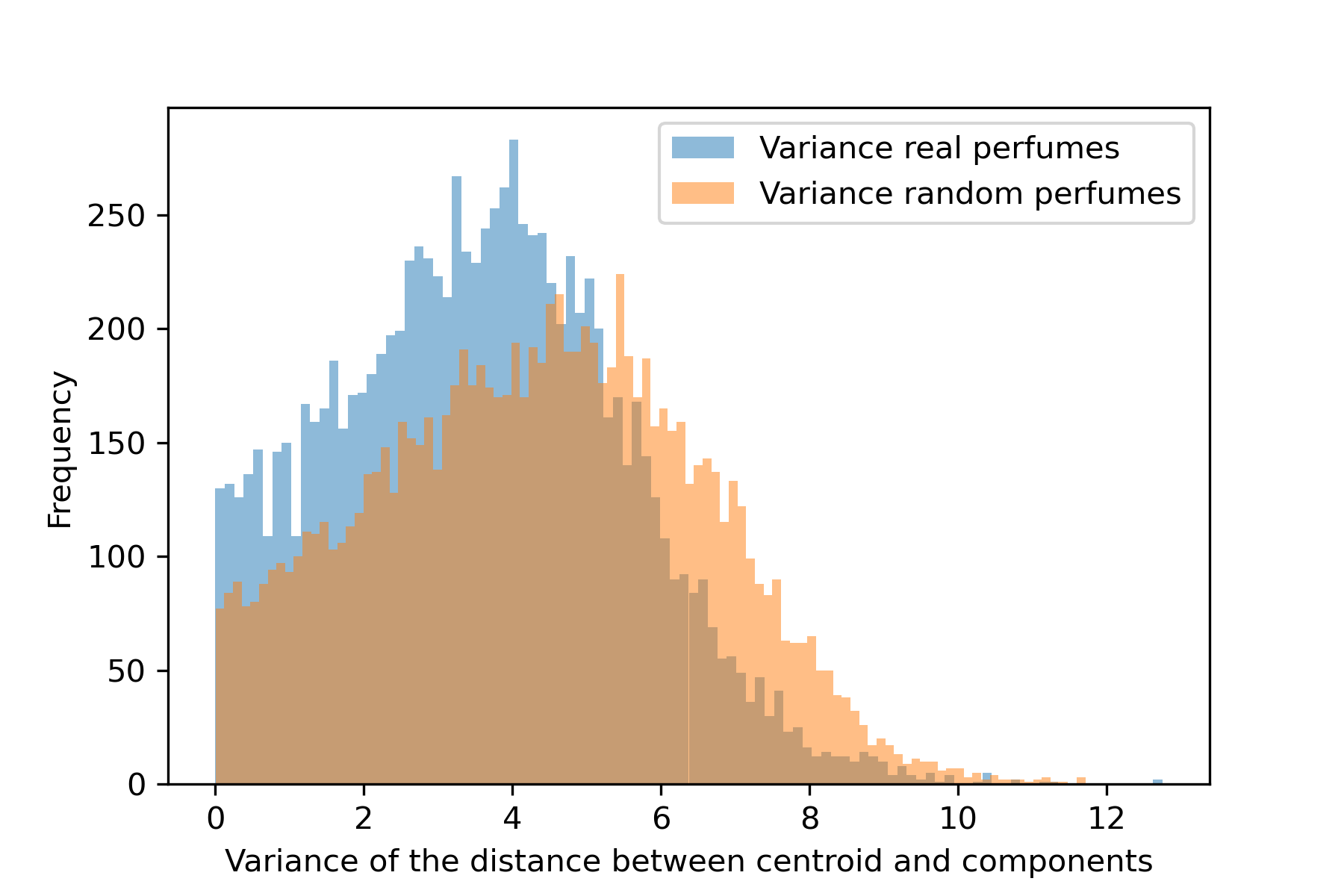} }}
    \qquad
    \subfloat[Kernel Density Estimation plot]{{\includegraphics[width=0.4\textwidth]{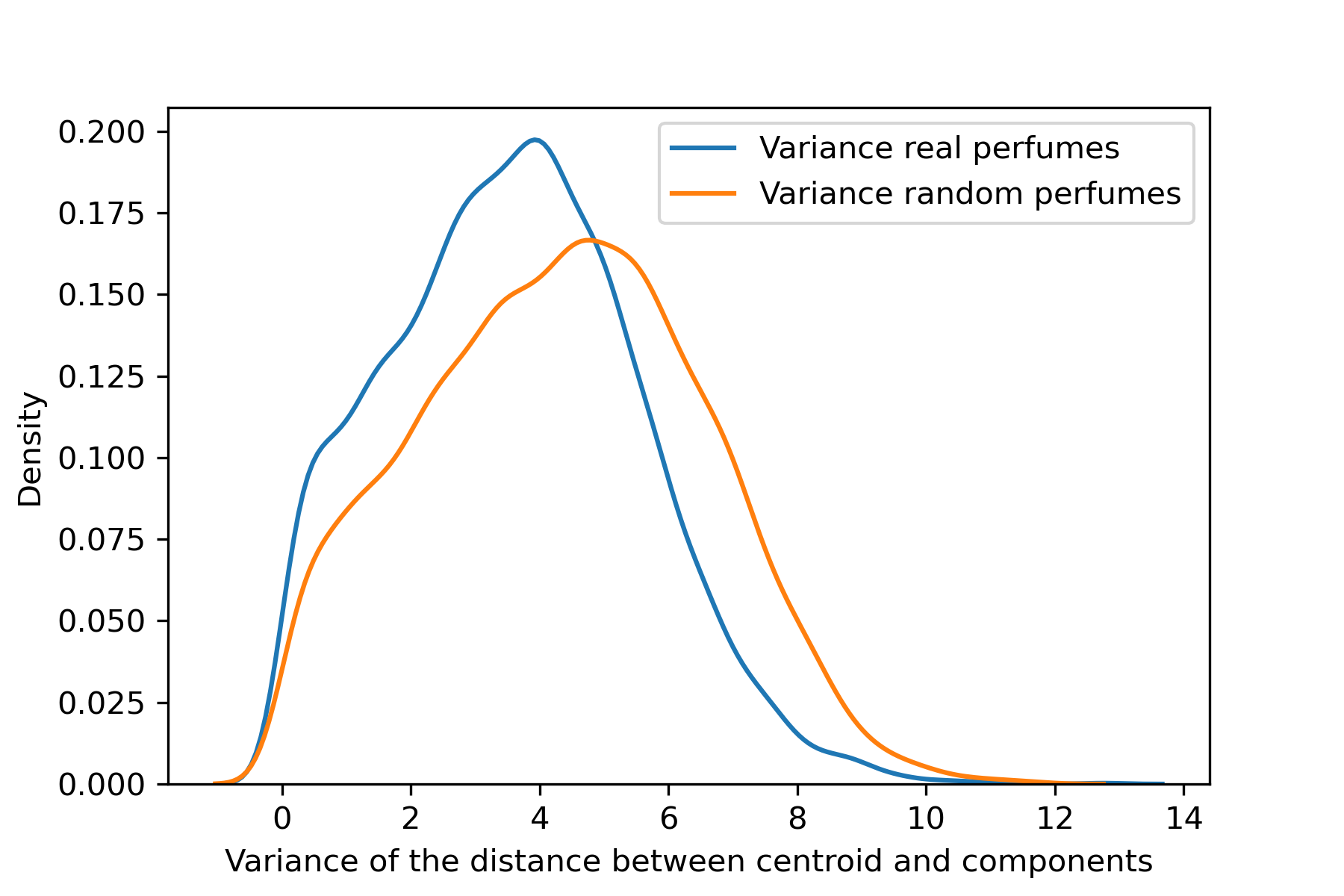} }}
    \caption{These two figures represent the average variance of the distance between the centroid of a perfume and its notes.}
    \label{fig:variance_diffs}
\end{figure}

In order to check whether this hypothesis is true, we calculated the centroid of each perfume\footnote{Similarly to \cite{morris2012soup} we these perfumes as being pleasant and of aesthetic value based on the fact they appear on this website.}, and then, the euclidean distance of each of the notes to the centroid. We then saved the variance for each perfume, as a measure of such variation. Besides, we generated a number of random perfumes\footnote{We made sure that the number of notes in these random perfumes was similar to the actual perfumes, and also that the random notes were drawn from the same distribution as the real perfumes.} and we did the same with their variances.

\subsubsection*{Results}
Figure \ref{fig:variance_diffs} shows a histogram and a Kernel Density Estimation plot to visualize the differences of the variances between real perfumes and randomly generated perfumes. We can see that there are some differences, and the difference in means is significant according to a Mann-Whitney-U Test ($p < 0.0001$). But, as it can be seen in both plots from Figure \ref{fig:variance_diffs}, the overlapping area between the two cases is quite large.


\section{Discussion}
In this section, we discuss the performed experiments and their results followed by an additional tentative experiment, where we explore the possibility of mappings from the word embedding space to the smell embedding space.
In experiment 1a, in general, we observed that the odor perception embeddings contain some information about smell. We evaluated this by calculating the Rank Biased Overlap (RBO) for each word in the shared vocabulary, effectively measuring the similarity between two rankings. One ranking was the one with the closest words from the smell embeddings (perfume) and the other ranking came from word embeddings (language). When comparing these RBOs with the ones obtained between smell embeddings and randomly shuffled word embeddings, we observed that there is a small but significant difference between the RBOs. This suggests, then, that there is an amount of shared information between the embedding spaces. However, the shared information seems to be rather limited which can be explained by the fact that general English does not express a large amount of olfactory information or the polysemy of the word embeddings (e.g. "orange" can be both a color and a smell, which probably increases the difference between "lemon" and "orange" in word embedding space).

Furthermore, these embeddings do not only seem to encode intuitive and established information, e.g. that \textit{lavender} and \textit{lavandin} are similar, but they also seem to capture subtle information, or information that may not be available at a first glance. For example, we found that \textit{musk} and \textit{jasmine} are relatively similar. While there are sources highlighting the "muskyness" of \textit{jasmine}, this characteristic is not reflected in traditional odor classification schemes.
However, when looking at the ranking based on closest words, we noticed a trend: The most similar notes also seem to be the most frequent ones in a number of cases.
We still need to analyze the effect of frequency in the embeddings, and we may need to account for that effect. A weighting scheme, similar to TF-IDF\footnote{Term-Frequency - Inverse Document Frequency.} could be applied as a possibility.

Concerning experiment 1b, the positive correlation between the RBOs and olfactory association indicates a tendency that the agreement between the spatial relationships between word embeddings and smell embeddings grows with the amount of olfactory information in a word embedding. This seems to support our assumption that at least a part of the little agreement can be explained by the general lack of olfactory information in English.

The goal of Experiment 2 was to get a first insight in whether the smell embeddings can potentially be used in order to model perfume aesthetics. Our results show a tendency that the notes of real perfumes are more evenly distributed around their centroid compared to randomly sampled perfumes. Although significant, the difference of the two groups was rather small, indicating that our measure only explains a small fraction of what makes perfumes pleasant. In general, these results encourage further research and analysis.

\subsection*{Further exploration}
After training different smell embedding models and evaluating them with Rank Biased Overlap, we decided to conduct another experiment. We tried to find a relation between the smell embedding space and the word embedding space. The final goal of this was to check whether we can find a sufficiently good mapping between two vectorial spaces and, if we could find one, whether we would be able to associate any word from a given language with a possible smell. This could potentially contribute to our understanding of how we associate words with smell.

For this experiment, the idea was to train a model that would receive an embedding table as input and tasking it with predicting the representations of each of those instances, but in the target space. We used the shared vocabulary of those two spaces as training data and we trained several regression models, specifically three different types of regression models: Linear Regression, Multilayer Perceptron Regressor and K-Nearest Neighbors Regressor\footnote{We use the \texttt{scikit-learn} \cite{pedregosa2011scikit} implementation of these models with the default parameters.}. We could have trained more types of regression models, but we decided to choose these because of the simplicity and in some cases, because of the ease of interpretation. For all these mapping experiments, we reduced the dimensionality of the word embeddings, from 300 dimensions to 20 dimensions, the same size of the odor perception embeddings. We used Principal Component Analysis (PCA) for this. The main reason behind this reduction was computational efficiency, as the number of learned parameters in the regressors would be significantly smaller. We validated these models using 5-Fold Cross-Validation. The obtained Mean-Squared Error (MSE) for each model can be found in table \ref{tab:regression_results_mse}. In addition, we also included a dummy regressor that constantly returns the mean value of the output, as a very simple baseline.

We obtained the best results with the MLP regressor model, but considering that Linear Regressor was relatively close, we decided to use the Linear Regressor for the final experiments. The main reason for choosing this model is the simplicity of the model, as it only performs a linear transformation to the input space, without further non-linearities, as it happens with the MLP regressor model.
It has to be noted here that the MLP regressor and the linear regressor perfomed only marginally better than the dummy regressor. A possible reason for this might be that these mappings are, if existent in the fist place, too complex for the chosen regression models. Alternatively, this might also be due to the fact the models are not sufficiently optimized, as this was beyond the scope of this experiment, or it might be a consequence of the dimensionality reduction.

In order to get some insight in how the linear regressor translates out-of-vocabulary words with no a priori associated smell, we showed the five most and least similar notes to some example words in Table \ref{tab:most_similar_mappings} together with some words that were part of the training set. As an additional reference, we also provide the most similar and least similar words according to the mappings estimated by the dummy regressor\footnote{As this baseline always returns the same value (mean), the most similar and least similar words are always the same for any input word.}: ``cypresses, balsa, currants, peppercorns, talcum'' were the most similar notes and ``carob, sake, sex, mamey, bean'' were the least similar for all cases. Admittedly, it seems hard to detect some general patters of how semantic qualities of the words are translated into smell, but looking at words such as "seduction" which produces a smell similar to \textit{pheromone} or "fish", which is translated to the smell \textit{kelp} (a seaweed), one gets the impression that there might be some semantic qualities that are captured by the mappings and translated in a way that makes intuitive sense. 
However, in general the results seem rather random. This might due to previously mentioned factors such as a potential non-linear relationship between the two embedding spaces, an insufficiently trained model or the reduced dimensions of the the input data. On top of this, polysemy in the word embeddings might add further complexity, e.g. "church" as organization vs. "church" as a building.
Especially concerning the latter, it might also be worth thinking about the word-smell pairs used for training, whose relationship is mostly source-based. Yet, most of us would probably associate "church" with \textit{incense} as opposed to \textit{stone}. Since this was only a tentative experiment, we did not investigate these issues any further, but aim to address them in future works.

\begin{table}[]
    \centering
\begin{tabular}{lrrrr}
\toprule
{} &  Linear Regression &  MLP Regression &  K-NN Regression &  Dummy Regression \\
\midrule
$\mu$    &           1.6473 &        1.6203 &         2.0472 &          1.7846 \\
$\sigma$ &           0.1383 &        0.1372 &         0.1895 &          0.1560 \\
\bottomrule
\end{tabular}
    \caption{Mean-Squared Error of different regression models in mapping from a source space (word embedding space) to a target space (smell embeddings space). The mean ($\mu$) and standard deviation ($\sigma$) were calculated from the 5-Fold Cross-Validation results.}
    \label{tab:regression_results_mse}
\end{table}


\begin{table}[]
    \centering
\begin{tabular}{lll}
\toprule
{} &                            5 most similar notes &                         5 least similar notes \\
word      &                                                 &                                                \\
\midrule
moon      &  cola, birch, raspberry, strawberry, sunflowers &     sulphur, buttercream, turf, fresh, lacquer \\
church    &            butter, earth, mugwort, stone, water &  grapefruit, sake, oleander, kumquat, cashmere \\
rotten    &     rubber, firewood, fireplace, fossils, cigar &     orchid, lemon, jasmine, bergamot, mandarin \\
seduction &         lys, leaf, gorse, broomstick, pheromone &  cinnamon, chocolate, fenugreek, coffee, carob \\
grass*     &        grass, rain, bulrush, thistle, blueberry &     sesame, mulberry, prune, elderberry, carob \\
night     &      sedum, papyrus, mesquite, sugar, honeycomb &   margarita, cassis, peonies, ozone, narcissus \\
wood*      &                wood, smoke, earth, fur, flowers &    eucalyptus, peonies, carob, sake, sandstone \\
male      &              scotch, kayak, ember, cigars, talc &    clover, aniseed, liquorice, rhubarb, acacia \\
female    &     heather, ember, turpentine, scotch, cypress &     rhubarb, pansy, aniseed, barley, liquorice \\
sweat*     &       saltwater, dust, chlorophyll, stalk, mist &    benzoin, cacao, verbena, cinnamon, mandarin \\
sky       &     foliage, port, parchment, lemonade, spirits &             rosewood, turf, mace, sulphur, ivy \\
cod       &          kirsch, algae, must, essence, pinewood &         champagne, cassia, cherry, carob, plum \\
fish      &           kelp, wheat, minerals, pandanus, peel &   cranberry, cinnamon, carob, papyrus, pimento \\
\bottomrule
\end{tabular}
    \caption{The 5 most similar and least similar notes to some selected words. The asterisk indicates that a word as pat of the training set.}
    \label{tab:most_similar_mappings}
\end{table}

\section{Conclusion and Future Work}
In this work, we used \textit{Word2vec} as a method for creating meaningful perfume note representations. We tested the quality of those embeddings by following a rank based metric (Rank Biased Overlap). Besides, experiment 1b suggests that the words with a higher RBO, or higher agreement in the ranking between note and word embedding spaces, are somehow more related to olfaction. Looking at the embeddings of particular notes, we discovered that they contain meaningful similarities to other notes that seem to capture both salient and more subtle similarities. Since these similarities sometimes transcend traditional categories, we believe that the embeddings presented in this paper could potentially contribute to new ways of odor categorization. Furthermore, our findings suggest that, in relation to distributional semantics, aesthetics in data with complex structures might be an equivalent concept to semantics in language. 

Additionally, we have performed an initial experiment in order to evaluate the potential use of our embeddings for modelling the pleasantness of a perfume. Our results suggest a tendency that the notes of real perfumes are more evenly distributed around their centroid compared to randomly sampled perfumes, but we believe this needs further analysis as the difference is relatively small. Since our results were nonetheless significant, we cautiously read them as encouragement for further exploration of geometric properties of perfumes in our embedding space in relation to their pleasantness.

To conclude, we explored the use of regression models for creating linear mappings between embedding spaces. This way, given a word in the English language, we can propose a perfume note based on this mapping. However, this approach needs further improvement.

In this work we did not associate each perfume with a specific gender, and we therefore considered all the perfumes' population as a whole. It would be very interesting to analyze the perfumes considering their gender association. Furthermore, it would also be relevant to investigate the grammatical gender of the odor descriptors, as done previously by Speed and others \cite{speed2019linguistic}. In order to do this, however, we would have to analyze the odor descriptors and their embeddings in languages other than English, in which words have an associated gender, such as German, Spanish, French, and so on.

With regards to preprocessing, there is considerable room for improvement: Applying lemmatization to all words might provide to be useful.
Additionally, we are planning on investigating the importance of the choice of hyperparameters of the word2vec algorithm in this setting, most notably the model architecture (CBOW vs. skip gram) and the size of the context window.


Furthermore, another possibility to improve our representations could be to include molecule information to these smell embeddings. In order to do so, we need a mapping between notes and molecules.

With regards to the linear mapping experiments, we believe that using more advanced mapping algorithms could make our results better and also that we would be able to provide more accurate odors for other non-odor related words. A possibility could be to follow the work on Unsupervised Machine Translation \cite{artetxe-etal-2018-unsupervised,lample2018unsupervised} and also Bilingual Lexical Induction \cite{artetxe2019bilingual}. In relation to this, we would like to have a proper and robust way for evaluating these mappings, besides qualitative and intuitive evaluations.

Last but not least, we want to continue trying to match different perception and language spaces. We have analyzed and mapped the spaces related to language and olfaction. Our intention is to do similar experiments with other sensory modalities, such as vision, taste or hearing.

\bibliography{references-mdk}
\end{document}